\documentclass[conference]{IEEEtran}
\IEEEoverridecommandlockouts
\usepackage{cite}
\usepackage{amsmath,amssymb,amsfonts}
\usepackage{algorithmic}
\usepackage{graphicx}
\usepackage{textcomp}
\usepackage{xcolor}

\usepackage{multirow}
\usepackage{makecell}
\usepackage{amsfonts}
\usepackage{enumerate}
\usepackage{stfloats}
\usepackage{float}
\usepackage{hyperref}
\usepackage{caption}
\usepackage{subcaption}

\hypersetup{
colorlinks=true,
linkcolor=black}

\def\BibTeX{{\rm B\kern-.05em{\sc i\kern-.025em b}\kern-.08em
    T\kern-.1667em\lower.7ex\hbox{E}\kern-.125emX}}
\begin{document}

\title{Continuous Prompt Tuning Based Textual Entailment Model for E-commerce Entity Typing
}

\author{\IEEEauthorblockN{Yibo Wang}
\IEEEauthorblockA{\textit{Department of Computer Science} \\
\textit{University of Illinois at Chicago}\\
Chicago, USA \\
ywang633@uic.edu}
\and
\IEEEauthorblockN{Congying Xia}
\IEEEauthorblockA{\textit{Saleforce Research}\\
USA \\
c.xia@salesforce.com}
\and
\IEEEauthorblockN{Guan Wang}
\IEEEauthorblockA{\textit{Huski Inc}\\
USA \\
guan.w@huski.ai}
\and
\IEEEauthorblockN{Philip S. Yu}
\IEEEauthorblockA{\textit{Department of Computer Science} \\
\textit{University of Illinois at Chicago}\\
Chicago, USA \\
psyu@uic.edu}
}

\maketitle

\begin{abstract}
The explosion of e-commerce has caused the need for processing and analysis of product titles, like entity typing in product titles. However, the rapid activity in e-commerce has led to the rapid emergence of new entities, which is difficult to be solved by general entity typing. Besides, product titles in e-commerce have very different language styles from text data in general domain. In order to handle new entities in product titles and address the special language styles problem of product titles in e-commerce domain, we propose our textual entailment model with continuous prompt tuning based hypotheses and fusion embeddings for e-commerce entity typing. First, we reformulate the entity typing task into a textual entailment problem to handle new entities that are not present during training. Second, we design a model to automatically generate textual entailment hypotheses using a continuous prompt tuning method, which can generate better textual entailment hypotheses without manual design. Third, we utilize the fusion embeddings of BERT embedding and CharacterBERT embedding with a two-layer MLP classifier to solve the problem that the language styles of product titles in e-commerce are different from that of general domain. To analyze the effect of each contribution, we compare the performance of entity typing and textual entailment model, and conduct ablation studies on continuous prompt tuning and fusion embeddings. We also evaluate the impact of different prompt template initialization for the continuous prompt tuning. We show our proposed model improves the average F1 score by around 2\% compared to the baseline BERT entity typing model.
\end{abstract}

\begin{IEEEkeywords}
continuous prompt tuning, textual entailment model, characterBERT, e-commerce entity typing
\end{IEEEkeywords}

\section{Introduction}
Nowadays, the boom of e-commerce has led to an increasing preference for online shopping. E-commerce platforms offer a wide variety of products. In order to better manage products and provide services to customers, such as classifying products and recommending products to customers, e-commerce platforms need to understand entities in the product titles, such as brand names (Apple, Nike, etc.), product names (iPhone, shoes, etc.) and other features (colors, sizes, etc.). This task is usually formulated as entity typing, which is to classify the entity types given the entity and context. For example, a product title `NAGANO Set of 2 Chairs' in Table \ref{tab:example data} has `NAGANO' as a brand name and `Chairs' as a product name. Thus the entity typing task is to classify `NAGANO' and `Chairs' into `brand name', `product name' or `feature' based on the entire product title.


Traditional approaches treat entity typing as a text classification problem with a pre-defined label set. 
However, new brands, products and features are emerging all the time on e-commerce platforms. For example, `iPhone 13 Pro Max' was released in September 2021 so the entity name `iPhone 13 Pro Max' will not appear in product titles before September 2021. Previous entity typing methods lack the flexibility to welcome new entities and require a large amount of training data to obtain decent performance. 

Thus, in this paper we propose a novel approach to formulate entity typing as a textual entailment problem. This textual entailment formulation gives our model the ability to easily adapt to new entities \cite{b1}. A textual entailment model is to determine whether two text fragments have a relationship in a sense that one fragment implies the other, that is, whether the two text fragments are entailment or not \cite{b2}. In our work, we formulate the textual entailment model by using the product title as one fragment of text and hypotheses with unfilled slots for entities to be classified as the other fragment of text. So the textual entailment task is to determine whether the product title and the hypotheses with entities to be classified are entailment or not. For example, when the product title is `Nike Mens Sportswear Aimoji Hoodie', the entity to be classified is `Hoodie' and entity labels are directly used as hypotheses, the textual entailment model input will be `Nike Mens Sportswear Aimoji Hoodie [SEP] Hoodie is a brand name/ product/ feature', where `is a brand name/ product/ feature' is the hypotheses.

\begin{table}
\caption{Sample data.}
\centering
\begin{tabular}{l}
\hline 
Product Titles$^{\mathrm{a}}$ \\
\hline
\textit{NAGANO} Set of 2 \textbf{Chairs} \\
\hline
\textit{Maison Louis Marie} \textbf{Perfume Oil Discovery Set} \\
\hline
\textit{Hugo Boss} (Dark Blue for Men 2.5 Oz) \textbf{EDT}\\
\hline
(100\% Cotton) Unpaper \textbf{Towels} - \textit{Cheeks Ahoy}\\
\hline
\multicolumn{1}{l}{$^{\mathrm{a}}$Italics indicates brand; Bold indicates product; Bracket indicates feature.}
\end{tabular}
\label{tab:example data}
\end{table}

\begin{table*}[h]
\caption{Examples of human designed hypotheses for our textual entailment models.}
\centering
\begin{tabular}{lll}
\hline 
Labels & \makecell[l]{Example Hypotheses \\Using Label Names} & \makecell[l]{Example Hypotheses Using\\ Label Dictionary Explanations}\\
\hline
Brand & `entity' is a brand name & \makecell[l]{`entity' is a band name, which is a type of things\\ manufactured by a particular company under a\\ particular name}\\
Product & `entity' is a product & \makecell[l]{`entity' is a product, which is an article or substance\\ that is manufactured or refined for sale} \\
Feature & `entity' is a feature & \makecell[l]{`entity' is a feature, which is a distinctive attribute \\or aspect of something}\\ 
\hline
\end{tabular}
\label{tab:example hypotheses}
\end{table*}

Designing the hypotheses is the most essential thing in the textual entailment task \cite{b1}.
Normally people will manually design multiple templates for experiments, and then choose the templates with best experimental performance.
Previous works usually convert labels or dictionary explanations into hypotheses with human-designed templates as shown in Table \ref{tab:example hypotheses}. 
However, the model performance cannot be guaranteed to be satisfactory using human designed hypotheses. Not to mention that it requires a lot of human effort and domain knowledge to design and test these templates.
Therefore, we propose a special continuous prompt tuning method on BERT to automatically find the best hypotheses for textual entailment models. Prompt tuning is the process of creating prompt templates (hypotheses in our case) that result in the most effective performance on the downstream tasks, and continuous prompt tuning is to directly create prompt templates in the embedding space rather than creating prompt templates that human can understand \cite{b3}. To the best of our knowledge, our work is the first to utilize continuous prompt tuning to design textual entailment hypotheses for entity typing.

Besides, product title data has its own language styles which are different from text data in general domain. Most product titles are not complete sentences, but phrases or a bunch of keywords which summarize and describe products. For example, `Trendy Apple Sports iWatch Band' is a phrase and `Zebra Sarasa Retractable Gel Pen | Blue | Medium 0.7 mm' is a set of keywords. Also, many special abbreviations like `Pcs' in `INGCO 142 Pcs combination tools set', self-made words like `Fashern' in `Fashern snake skin print rebecca set' and translated words like `GaGaZui' in `GaGaZui Green Bean Sweet Spicy Flavor 33g * 30 bags 990g' are in product titles. In order to adjust to these e-commerce domain special language styles, we utilize fusion embeddings of BERT and CharacterBERT \cite{b4}, which is a character-level language model that includes a Character-CNN module to represent entire words by deliberating their characters rather than using predefined word-piece vocabularies from general domain as in BERT \cite{b5}. By using fusion embeddings of BERT embedding and CharacterBERT embedding, the special language styles of product titles and the tokens that do not appear in the predefined word-piece vocabularies can be handled better.

In this paper, we propose \textbf{C}ontinuous Prompt Tuning based \textbf{T}extual Entailment \textbf{M}odel (CTM) to handle new entities in product titles, automatically generate optimized textual entailment hypotheses and address the special language styles problem of e-commerce domain. Our primary contributions include:
\begin{itemize}
    \item Formulating entity typing in product titles into a novel textual entailment problem and building the textual entailment model hypotheses using continuous prompt tuning methods: The textual entailment formulation provides model with the ability to handle new entities that are not present in training set.
    \item Building a continuous prompt tuning model for BERT and characterBERT: The continuous prompt tuning model saves human effort to design textual entailment hypotheses and is able to automatically build better hypotheses.
    \item Utilizing the fusion embeddings of BERT embedding and CharacterBERT embedding: The fusion embeddings of BERT embedding and CharacterBERT embedding allow our proposed CTM model to handle the special language styles of e-commerce product titles.
\end{itemize}

The paper is structured as follows: in Section \ref{sec:2} we introduce recent works on applications of textual entailment model, different types of prompt tuning methods and character-level language models; in Section \ref{sec:3} we describe our proposed CTM model in detail; in Section \ref{sec:4} we illustrate our dataset and experimental details and analyze the experimental results; in Section \ref{sec:5} we summarize the contents of our paper.

Our codes are available at \href{https://github.com/YiboWANG214/CTM}{GitHub link}.

\section{Related Work}
\label{sec:2}
Textual entailment problem is a frequently studied task, and both prompt learning and character-level language models are very popular in recent years.

Textual entailment model was first proposed as a natural language processing task by \cite{b6} in 2005. Recently, lots of work has been studied for converting a classification task into a textual entailment problem. For example, \cite{b7} converts a relation classification task into a textual entailment problem, where the hypotheses are relation descriptions. \cite{b1} treats the zero shot text classification task as a textual entailment problem, so that their model can achieve knowledge from other entailment datasets. \cite{b8} frames the fine-grained classification of socio-political events into a textual entailment problem, whose task is that given a text and a event class, the model should decide whether the text describes the event of the given class. The textual entailment problem has great potential because the changes and design of hypotheses can bring about many model modifications and improvements. In our work, we convert an entity typing task into a textual entailment problem to deal with new entities that are not present in training data and better exploit contextual information and textual entailment hypotheses.

Since the publication and popularity of GPT-3 \cite{b9}, prompt learning has been payed great attention to. As mentioned in \cite{b3}, prompt tuning is the phase of creating effective prompt templates for downstream tasks, which is one of the most important phases of prompt learning. 
Prompt tuning can be classified into hand-crafted prompt tuning and automated prompt tuning. Many famous work has exploited hand-crafted prompt tuning, like GPT-3 \cite{b9}, T5 \cite{b10}, LAMA \cite{b11}, CTRL \cite{b12}, etc. Manually designed prompt templates do not require computational resources and work well in many cases. However, exploring optimal prompt templates via hand-crafted prompt tuning is hard and requires time and domain knowledge. Thus, automated prompt tuning is proposed to solve these problems. Automated prompt tuning can be further classified into discrete prompt tuning and continuous prompt tuning. Discrete prompt tuning is to automatically design prompt templates in discrete space that are natural languages human can understand. For example, AutoPrompt \cite{b13} and AdvTrigger \cite{b14} are methods based on a gradient-guided search to automatically create discrete prompt templates; \cite{b15} proposes mining-based and paraphrasing-based methods to automatically generate discrete prompt templates. Continuous prompt tuning (also known as soft-prompt tuning) is to create prompt templates in embedding space. For example, WARP \cite{b16} is a method based on adversarial reprogramming to learn task-specific word embeddings and create continuous prompt templates; prefix-tuning \cite{b17} builds continuous prompts by tuning a small continuous task-specific vector and keeping language model parameters frozen; soft-prompt tuning \cite{b18} is a simplification of prefix-tuning which directly tunes the embeddings of the prepended tokens. There are many other excellent prompt tuning works, such as \cite{b19, b20, b21, b22, b23, b24, b25, b26, b27, b28, b29} and so on. Prompt learning has an effect comparable to or even better than fine-tuning on large language models such as GPT-3 and T5 \cite{b17, b18} and smaller language models like \cite{b30}. 
The success of prompt learning inspires us to apply the idea of continuous prompt tuning to design textual entailment hypotheses for entity typing task. It should be noted that most of the previous research on prompt learning investigate for zero-shot/few-shot experiments, while only a few studies focus on generating templates using prompt tuning for other tasks.
CONAN \cite{b31} uses continuous patterns for textual entailment model by using a set of fresh tokens to separate the two text fragments in textual entailment models. In CONAN, the fresh tokens are randomly initialized without any further training or tuning, while in our work our hypotheses are tuned to optimal. \cite{b32} applies prompt learning to entity typing, but they use a more typical prompt learning pipeline, which includes label words construction and prompt templates design. In \cite{b32}, they use confidence scores of all the label words to construct the final score of the particular entity type in mask language modeling.
In our work, we first use continuous prompt tuning to design optimal textual entailment hypotheses. After obtaining the optimized textual entailment hypotheses, we apply these hypotheses to a textual entailment model, which is reformulated from entity typing task, using a fine-tuning paradigm. 

Many character-level language models based on different methods have been proposed to tackle the problem caused by subword-level models. \cite{b33} introduces a hierarchical RNN based character-level language model. \cite{b34} proposes a model to apply a CNN and a highway network over characters, whose output is sent to a LSTM lanaguage model. CharacterBERT \cite{b4} utilizes a Character-CNN module to represent tokens by their characters; CharBERT \cite{b35} uses context string embeddings and a heterogeneous interaction module to obtain character representations; CharFormer \cite{b36} use a soft gradient-based subword tokenization module to learn subword representation from characters. Canine \cite{b37} is a neural encoder that operates directly on character sequences without subword tokenization or vocabulary. We utilize the fusion embedding of BERT embedding and CharacterBERT embedding to address the special language styles of e-commerce product titles.

\section{Methods}
\label{sec:3}
In this work, we reformulate the entity typing task in product titles into a textual entailment problem. To save human effort and improve model performance, we utilize continuous prompt tuning to automatically create optimized hypotheses for the textual entailment model. Additionally, we fuse embeddings of BERT embedding and characterBERT embedding to acclimate to special language styles of product titles. 

\begin{figure*}
	\centering
	\includegraphics[width=1\textwidth]{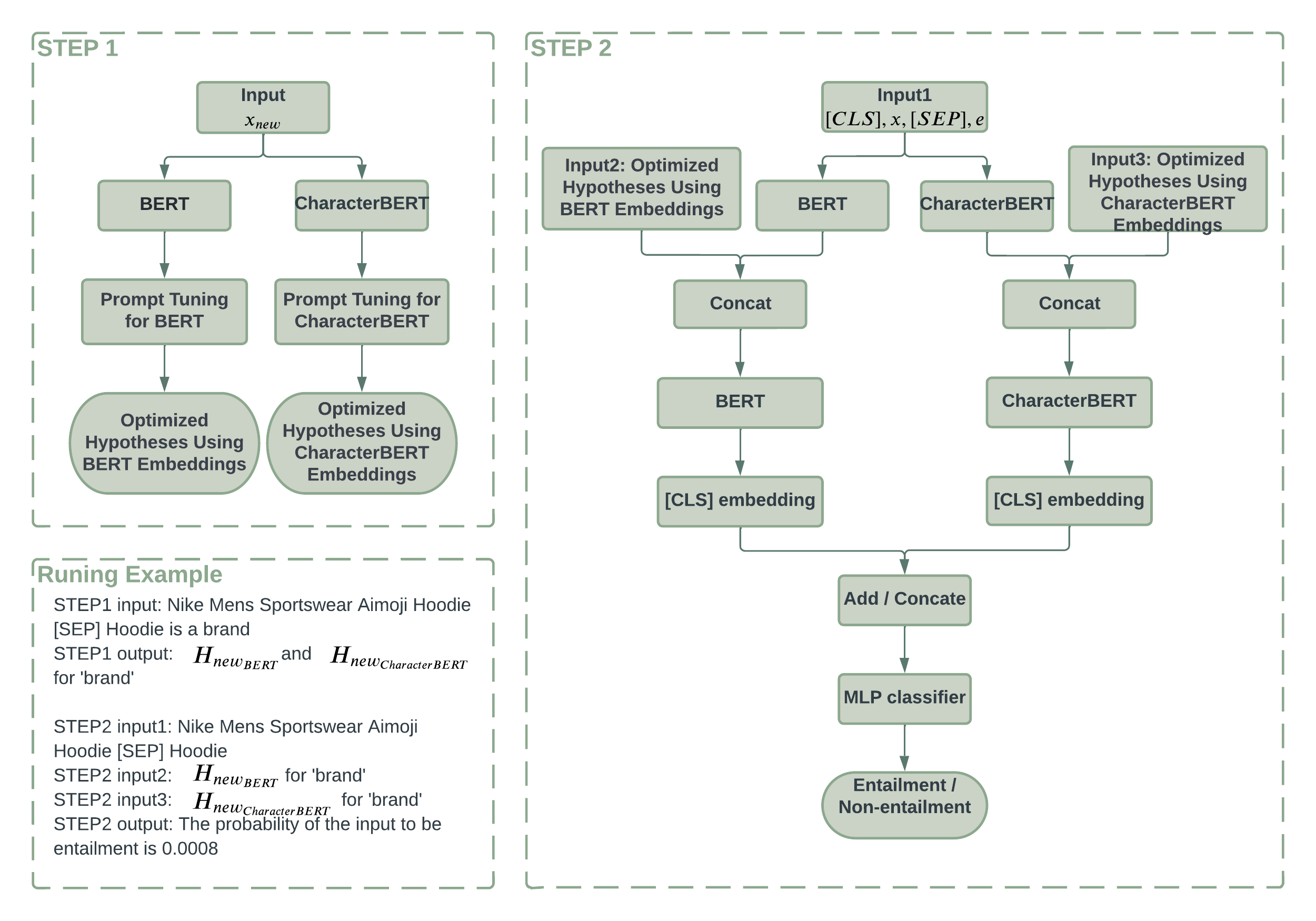}
	\caption{Model architecture. STEP 1 is the process of obtaining optimized hypotheses by applying prompt tuning to BERT and CharacterBERT separately, and STEP 2 is the process of using textual entailment model and fusion embedding for classification. A running example is shown in the figure.} 
	\label{fig:1}
\end{figure*}

In this section, we first give a problem formulation of the entity typing task and the textual entailment task (\ref{subsec:1}), followed by details of our model architecture, including textual entailment model (\ref{subsec:2}), continuous prompt tuning (\ref{subsec:3}) and fusion embedding (\ref{subsec:4}). The model architecture is shown in Fig. \ref{fig:1}.

\subsection{Problem Formulation}
\label{subsec:1}
Given a product title $x$ and an entity $e$ which is a substring of $x$, the original entity typing task is to classify the entity $e$ into an entity type $y \in \{\text{`brand', `product', `feature'}\}$ for each input product title $x$. For example, in the product title 'Nike Mens Sportswear Aimoji Hoodie', the entity 'Hoodie' should be classified as product rather than brand name or features.

In order to format the entity typing task into a textual entailment problem, we concatenate the product title $x$, the entity $e$ and the hypothesis $h$ for different classes together and send them into the textual entailment model. 
The new input to the textual entailment model is formulated as follows:
\begin{equation}
    x_{new} = [CLS], x, [SEP], e, h, \label{eq:model_input}
\end{equation}
where $h \in \mathbb{H}$, and
$\mathbb{H}$ is the hypotheses set with one hypothesis for each class.
The objective of the textual entailment model is to predict whether the two text fragments of the concatenated new input is entailment or not, which is to classify the new input $x_{new}$ into a binary class $y_{new} \in \{\text{`entailment', `non-entailment'}\}$.

\subsection{Textual Entailment model}
\label{subsec:2}
In order to train a textual entailment model, we need to construct positive/negative entailment pairs. For the positive pairs, we concatenate the sentence-entity pair and the hypothesis from its ground truth label as the positive entailment example.
For negative pairs, we randomly select one hypothesis from negative labels to construct `non-entailment' example.
For example, `Nike Mens Sportswear Aimoji Hoodie [SEP] Hoodie is a product’ is an entailment example, while `Nike Mens Sportswear Aimoji Hoodie [SEP] Hoodie is a brand name’ and `Nike Mens Sportswear Aimoji Hoodie [SEP] Hoodie is a feature’ are non-entailment examples.

During training, we train the textual entailment model with the cross entropy loss on the constructed positive and negative entailment pairs.
For the inference, we concatenate each test example with hypothesis from different classes and we choose the class with the highest probability as the predicted entity type.

\subsection{Continuous Prompt Tuning}
\label{subsec:3}
Inspired by soft prompt tuning \cite{b18}, 
we design a novel continuous prompt tuning method to obtain optimized hypotheses for the textual entailment task.

Given a product title with $n$ tokens $x=\{x_1, \dots , x_n\}$ and the corresponding entity with $m$ tokens $e = \{e_1, \dots , e_m\}$, the embedding matrix $X_e = [X;E] \in \mathbb{R}^{(n+m)\times d}$ is obtained by the language model, where $X \in \mathbb{R}^{n \times d}$ is the embedding matrix of the product title $x$, $E \in \mathbb{R}^{m \times d}$ is the embedding matrix of the entity $e$ and $d$ is the dimension of the embedding space. The continuous prompt template (hypothesis) $h$ is represented as $H \in \mathbb{R} ^ {p \times d}$, where $p$ is the length of the prompt tokens. Then the embedding matrix $X_e$ and the prompt matrix $H$ are concatenated as $[X_e; H] \in \mathbb{R} ^{(n+m+p) \times d}$ to be the input embedding of the language model. The prompt tuning phase is then modeled as a masked language modeling. During the fine-tuning of the masked language modeling, the parameters $\theta$ of the language model are frozen, and the only tunable parameters are the embedding matrix $H$ of the hypothesis. For each class, the optimized embedding matrix $H_{new} \in \mathbb{R}^{p\times d}$ of one hypothesis is generated by using all training data in this class. Later these embedding matrices will be used as the optimized hypotheses in the textual entailment model. 

The initialization of the hypotheses is critical to the continuous prompt tuning and the textual entailment model performance. We use different lengths and different types of the hypotheses initialization. The first type of hypotheses is directly using class labels to formulate the initialized hypotheses. The length of the hypotheses is 3 and the hypotheses are `is a brand/ is a product/ is a feature'. The second type of hypotheses is converted from dictionary explanations of the class labels. The length of the hypotheses is 14 and the initialized hypotheses are `is a brand, which is a type of things manufactured by a particular company/ is a product, which is an article or substance manufactured or refined for sale/ is a feature, which is a distinctive attribute or a special aspect of something'. Different initialization of the hypotheses leads to different model performance.

\subsection{Fusion Embeddings}
\label{subsec:4}
For CharacterBERT, each token is converted into a sequence of characters and then represented as a 50-dimension input tensor. This 50-dimension input tensor can be considered similar to input id in BERT framework. Then a Character-CNN module will be used to project the 50-dimension input tensor into a 768-dimension CharacterBERT embedding representation which is selected to be dimensionally aligned with BERT.

Although BERT and CharacterBERT use different ways to process input, they can handle the same input in their own way and obtain output of the same dimension. Thus, the optimized hypothesis using BERT embedding $H_{new_{BERT}}$ and the optimized hypothesis using CharacterBERT embedding $H_{new_{CharacterBERT}}$ for each class can be obtained by applying continuous prompt tuning separately on BERT model and CharacterBERT model. Then $H_{new_{BERT}}$ should be concatenated to the BERT embedding of the product title and the entity, and the concatenated new BERT embedding should be fed into the BERT model for a better contextual embedding. Similarly, $H_{new_{CharacterBERT}}$ should be concatenated to the CharacterBERT embedding of the product title and the entity, and the concatenated new CharacterBERT embedding should be fed into the CharacterBERT model.

After obtaining the better contextual BERT embedding and CharacterBERT embedding, we use different fusion methods to obtain the fusion embedding of BERT embedding and CharacterBERT embedding.
The same embedding dimension of CharacterBERT and BERT allows diverse fusion methods to be tried. We tried two simple but effective fusion methods: 1). concatenating `[CLS]' BERT embedding ($V_{BERT}$) and weighted `[CLS]' CharacterBERT embedding ($V_{CharacterBERT}$), and 2). adding `[CLS]' BERT embedding ($V_{BERT}$) to weighed `[CLS]' CharacterBERT embedding ($V_{CharacterBERT}$). We use `[CLS]' embedding because `[CLS]' embedding is considered to accommodate sentence information. 

For the first fusion method, the fusion embedding for one sample is
\begin{equation}
    [V_{BERT}\quad \alpha * V_{CharacterBERT}],
\end{equation}
where $V_{BERT}$ is `[CLS]' BERT embedding, $V_{CharacterBERT}$ is `[CLS]' CharacterBERT embedding and $\alpha$ is a tunable weight parameter. The fusion embedding shape for a batch is [batch size, 2*hidden\_size]. For the second method, the fusion embedding for one sample is
\begin{equation}
    V_{BERT} + \alpha * V_{CharacterBERT}.
\end{equation}
The fusion embedding shape for a batch is [batch size, hidden\_size]. The fusion embeddings of input are then fed into two-layer MLPs for textual entailment classification.

In summary, as in Fig. \ref{fig:1}, the input of STEP 1 is $x_{new}$ in (\ref{eq:model_input}), which is the new input for the textual entailment formulation and the output of STEP 1 is the optimized hypothesis using BERT embedding $H_{new_{BERT}}$ and the optimized hypothesis using CharacterBERT embedding $H_{new_{CharacterBERT}}$. The input of STEP 2 is $x$, $e$, $H_{new_{BERT}}$ and $H_{new_{CharacterBERT}}$, and the output of STEP 2 is the probability of the input to be entailment.

\begin{table*}[h]
\caption{Hyperparameters for entity typing models.}
\centering
\begin{tabular}{llllllll}
\hline 
Model & Dataset & max\_seq & batch\_size & learning\_rate & epoch & $\alpha$\\
\hline
Entity Typing & test & 64 & 32 & 5e-6 & 7 & -\\
Entity Typing & novel entity & 64 & 32 & 5e-6 & 10 & -\\
\hline
\end{tabular}
\label{tab: hyperparameters1}
\end{table*}

\begin{table*}
\caption{Hyperparameters for models using labels as hypotheses or prompt initialization.}
\centering
\begin{tabular}{llllllll}
\hline 
Model & Dataset & max\_seq & batch\_size & learning\_rate & epoch & $\alpha$\\
\hline
Textual Entailment & test & 64 & 16 & 5e-6 & 6 & -\\
Textual Entailment & novel entity & 64 & 16 & 3e-6 & 7 & -\\
\hline
CTM (add) & test & 64 & 64 & 1e-5 & 4 & 1.0072\\
\ (w/o) prompt tuning & test & 64 & 16 & 1e-5 & 10 & 1.0099\\
CTM (concate) & test & 32 & 32 & 1e-5 & 7 & 10068\\
\ (w/o) prompt tuning & test & 64 & 32 & 3e-6 & 10 & 1.0044\\
CTM (CharacterBERT only) & test & 64 & 16 & 8e-6 & 9 & -\\
CTM (BERT only) & test &  64 & 16 & 8e-6 & 10 & -\\
\hline
\end{tabular}
\label{tab: hyperparameters2}
\end{table*}

\begin{table*}
\caption{Hyperparameters for models using dictionary explanations as hypotheses or prompt initialization.}
\centering
\begin{tabular}{llllllll}
\hline 
Model & Dataset & max\_seq & batch\_size & learning\_rate & epoch & $\alpha$\\
\hline
Textual Entailment & test & 64 & 16 & 3e-6 & 8 & -\\
Textual Entailment & novel entity & 64 & 16 & 3e-6 & 9 & -\\
\hline
CTM (add) & test & 64 & 4 & 1e-5 & 10 & 1.0083\\
\ (w/o) prompt tuning & test & 64 & 16 & 1e-5 & 3 & 1.0047\\
CTM (concate) & test & 64 & 8 & 1e-5 & 5 & 1.0063\\
\ (w/o) prompt tuning & test & 64 & 16 & 5e-6 & 3 & 1.0039\\
CTM (CharacterBERT only) & test & 32 & 16 & 5e-6 & 6 & -\\
CTM (BERT only) & test & 64 & 16 & 8e-6 & 10 & -\\
\hline
\end{tabular}
\label{tab: hyperparameters3}
\end{table*}

\section{Experiments and Results}
\label{sec:4}
In this section, we compare the performance of our proposed CTM model and the baseline models on a product title dataset. Then we compare model performance of entity typing and textual entailment model on test set and novel entity set and conduct a series of ablation studies to investigate the effect of each contribution. Through the experiments and performance comparison, we aim to answer the following research questions regarding our proposed continuous prompt tuning based textual entailment model:
\begin{itemize}
    \item \textbf{RQ1.} Does the textual entailment formulation have better ability to handle new entities compared with the original entity typing task?
    \item \textbf{RQ2.} Does the different hypotheses selection affect the performance of the textual entailment model?
    \item \textbf{RQ3.} Does the hypotheses obtained by prompt tuning work better than hand-crafted hypotheses? Do different initialized hypotheses for prompt tuning have different effect on model performance?
    \item \textbf{RQ4.} Does using the fusion embedding work better than using BERT embedding alone or using CharacterBERT embedding alone?
\end{itemize}

\subsection{Dataset}
\label{sec:4.1}
The raw data is provided by Huski.ai \footnote{https://huski.ai}.
We randomly select 10,000 English product titles from the raw data, and hire three workers to label each token in product titles as one of the brand name, product/ product name, feature or others. We then filter the data by following steps: 1) reformulate data into a `entity type, entity, product title' format for each annotator; 2) preserve samples that at least two annotators agree with and discard other samples; 3) manually double check whether the annotations of the preserved samples conform to common sense and discard clearly mislabeled data. After filtering, the training set has 2324 samples in total with 797 `brand', 700 `product' and 827 `feature', and the test set has 1317 samples in total with 360 `brand', 437 `product' and 520 `feature'. We also create a novel entity set, which is filtered from the test set, only with new entities that are not present in training set. The novel entity set has 572 samples in total with 267 `brand', 135 `product' and 170 `feature'. This novel entity set is used to compare the ability to handle new entities between the entity typing task and the textual entailment formulation. Sample data is shown in Table \ref{tab:example data}.

\subsection{Experimental Setup}
\label{sec:4.2}
All the experiments are conducted on one Tesla P100 GPU. The models are based on BERT or CharacterBERT, whose hidden\_size is 768. 
We use two-layer MLPs for classification, with ReLU activation functions and the size of each hidden layer being 100 and 50. For our proposed CTM model, when prompt initialization is converted from class labels and using addition for fusion embeddings, $\alpha = 1.0072$, $max\_seq=64$, $batch\_size=64$, $learning\_rate=1e-5$, $epoch=4$; when prompt initialization is converted from class labels and using concatenation for fusion embeddings, $\alpha = 1.0068$, $max\_seq=32$, $batch\_size=32$, $learning\_rate=1e-5$, $epoch=7$; when prompt initialization is converted from dictionary explanations and using addition for fusion embeddings, $\alpha = 1.0083$, $max\_seq=64$, $batch\_size=4$, $learning\_rate=1e-5$, $epoch=10$; when prompt initialization is converted from dictionary explanations and using concatenation for fusion embeddings, $\alpha = 1.0063$, $max\_seq=64$, $batch\_size=8$, $learning\_rate=1e-5$, $epoch=5$. Hyperparameters for all the models are reported in Table \ref{tab: hyperparameters1}, Table \ref{tab: hyperparameters2} and Table \ref{tab: hyperparameters3}.

\begin{table*}
\caption{F1 score comparison between our proposed CTM models and the baseline model.}
\centering
\begin{tabular}{llllll}
\hline 
Hypotheses / Initialization & Model & Brand & Product & Feature & Average \\
\hline 
- & Entity Typing & 0.8602 & 0.8391 & 0.8754 & 0.8595\\
\hline
\multirow{2}*{Class Labels} & CTM (add) & \textbf{0.8809} & 0.8555 & 0.8931 & 0.8770 \\
~ & CTM (concate) & 0.8776 & 0.8587 & 0.8871 & 0.8747\\
\hline
\multirow{2}*{Dictionary Explanations} & CTM (add) & 0.8760 & 0.8636 & 0.8906 & \textbf{0.8778}\\
~ & CTM (concate) & 0.8679 & \textbf{0.8666} & \textbf{0.8946} & \textbf{0.8778} \\
\hline
\end{tabular}
\label{tab:general}
\end{table*}

\begin{table*}
\caption{Class F1 scores and Average F1 scores for entity typing and textual entailment model with labels / dictionary explanations as hypotheses on test set and novel entity set.}
\centering
\begin{tabular}{llllll}
\hline 
Dataset & Model & Brand & Product & Feature & Average \\
\hline
\multirow{3}*{Test Set} & Entity Typing & 0.8602 & 0.8391 & \textbf{0.8754} & 0.8595\\
~ & Textual Entailment (label) & 0.8568 & 0.8395 & 0.8605 & 0.8572\\
~ & Textual Entailment (dictionary) & \textbf{0.8776} & \textbf{0.8451} & 0.8727 & \textbf{0.8648}\\
\hline
\multirow{3}*{\makecell[c]{Novel Entity Set}} & Entity Typing & 0.8925 & 0.7860 & 0.8632 & 0.8601 \\
~ & Textual Entailment (label) & 0.9094 & 0.7833 & 0.8693 & 0.8689\\
~ & Textual Entailment (dictionary) & \textbf{0.9139} & \textbf{0.7985} & \textbf{0.8772} & \textbf{0.8759}\\
\hline
\end{tabular}
\label{tab:experimental results entailment}
\end{table*}

\begin{table*}
\caption{Class F1 scores and Average F1 scores for CTM and CTM without prompt tuning to evaluate the contribution of prompt tuning. All the results are evaluated on the test set.}
\centering
\begin{tabular}{llllll}
\hline 
Hypotheses / Initialization & Model & Brand & Product & Feature & Average \\
\hline
\multirow{4}*{Class Labels} & CTM (add) & \textbf{0.8809} & 0.8555 & \textbf{0.8931} & \textbf{0.8770} \\
~ & \ (w/o) prompt tuning & 0.8808 & \textbf{0.8575} & 0.8836 & 0.8740\\
~ & CTM (concate) & 0.8776 & \textbf{0.8587} & \textbf{0.8871} & \textbf{0.8747}\\
~ & \ (w/o) prompt tuning & \textbf{0.8862} & 0.8532 & 0.8749 & 0.8709\\
\hline
\multirow{4}*{Dictionary Explanations} & CTM (add) & \textbf{0.8760} & \textbf{0.8636} & \textbf{0.8906} & \textbf{0.8778}\\
~ & \ (w/o) prompt tuning & 0.8717 & 0.8523 & 0.8787 & 0.8679\\
~ & CTM (concate) & \textbf{0.8679} & \textbf{0.8666} & \textbf{0.8946} & \textbf{0.8778} \\
~ & \ (w/o) prompt tuning & 0.8668 & 0.8477 & 0.8784 & 0.8648\\
\hline
\end{tabular}
\label{tab:experimental results}
\end{table*}

\subsection{Experimental Results and Analysis}
\label{sec:4.3}
Table \ref{tab:general} exhibits the F1 score performance comparison between our proposed CTM models with different fusion methods and prompt initialization and the baseline entity typing model. Any version of our CTM model has better model performance compared to the entity typing model, and CTM with concatenation as fusion method and dictionary explanations as prompt initialization has the best performance in general.

To analyze the contributions of different components in our proposed CTM model, we compare model performance of entity typing and textual entailment model on test set and novel entity set, and perform ablation studies on prompt tuning and fusion embedding on test set. We show class F1 scores and average F1 scores of the models in Tables \ref{tab:general}, \ref{tab:experimental results entailment}, \ref{tab:experimental results} and \ref{tab:experimental results fusion embedding}.


\begin{table*}
\caption{Class F1 scores and Average F1 scores for CTM and CTM without fusion embedding to evaluate the contribution of fusion embedding. All the results are evaluated on the test set.}
\centering
\begin{tabular}{llllll}
\hline 
Hypotheses / Initialization & Model & Brand & Product & Feature & Average \\
\hline
\multirow{4}*{Class Labels} & CTM (add) & \textbf{0.8809} & 0.8555 & \textbf{0.8931}  & \textbf{0.8770}\\
~ & CTM (concate) & 0.8776 & \textbf{0.8587} & 0.8871 & 0.8747\\
~ & CTM (CharacterBERT only) & 0.8683 & 0.8459 & 0.8751 & 0.8633\\
~ & CTM (BERT only) & 0.8607 & 0.8368 & 0.8799 & 0.8610\\
\hline
\multirow{4}*{Dictionary Explanations} & CTM (add)  & 0.8760 & 0.8636 & 0.8906 & \textbf{0.8778}\\
~ & CTM (concate)  & 0.8679 & \textbf{0.8666} & \textbf{0.8946} & \textbf{0.8778}\\
~ & CTM (CharacterBERT only)& 0.8667 & 0.8389 & 0.8776 & 0.8618 \\
~ & CTM (BERT only)& \textbf{0.8850} & 0.8492 & 0.8809 & 0.8717 \\
\hline
\end{tabular}
\label{tab:experimental results fusion embedding}
\end{table*}


\noindent\textbf{Textual Entailment Formulation (RQ1, RQ2)}
In order to compare the ability to handle new entities between entity typing formulation and textual entailment formulation, we conduct comparative experiments between entity typing formulation and textual entailment formulation on both test set and novel entity set. Besides, we also conduct experiments on textual entailment model with different hand-crafted hypotheses to verify the different impact of different hypotheses.
The regular entity typing model is used as a baseline model, and BERT is used as a backbone model for both entity typing and textual entailment model.

As in Table \ref{tab:experimental results entailment}, on test set the average F1 score of the textual entailment reformulation with labels as hypotheses is 0.23\% lower than the average F1 score of the general entity typing task and the average F1 score of the textual entailment model with dictionary explanations as hypotheses is 0.53\% higher than that of the general entity typing; on the novel entity set the average F1 score of the textual entailment models with labels as hypotheses is 0.88\% higher than the average F1 score of the general entity typing task and the average F1 score of the textual entailment model with dictionary explanations as hypotheses is 1.58\% higher than that of the general entity typing. 
For test set, the difference of model performance between entity typing and textual entailment is relatively small and the average F1 score of entity typing is even slightly better than textual entailment model with labels as hypotheses. For novel entity set, the model performance improvement obtained by textual entailment formulation is more significant for hypotheses converted from both labels and dictionary explanations. For both sets, the average F1 scores of the textual entailment model with dictionary explanations as hypotheses are higher than that of the textual entailment model with labels as hypotheses.

The more obvious improvement of textual entailment model on novel entity set compared to the improvement of textual entailment model on test set shows that the textual entailment formulation provides model with the ability to better handle new entities. Beside, the format of textual entailment model provides more possibilities for model performance improvement, like the selection of hypotheses. The model performance difference between two hypotheses indicates that different textual entailment hypotheses lead to different model performance, which confirms our previous thinking that model performance cannot always be guaranteed by human-designed hypotheses and indicates the importance of the selection of hypotheses. 
The textual entailment formulation makes it possible to apply prompt tuning techniques to automatically create optimized hypotheses to improve model performance.

\noindent\textbf{Prompt Tuning (RQ3)} 
In order to evaluate the contribution of prompt tuning and find out whether the hypotheses obtained by prompt tuning work better than hand-crafted hypotheses, we compare the model performance between our proposed CTM models and the similar models without prompt tuning based hypotheses. Besides, to explore the effect of different prompt initialization for prompt tuning, the experiments are performed on hypotheses and prompt initialization converted from both labels and dictionary explanations.

For prompt tuning, as in Table \ref{tab:experimental results} when class labels are used for textual entailment hypotheses and prompt initialization, prompt tuning helps increase average F1 score by 0.30\% using addition as fusion method and increase average F1 score by 0.38\% using concatenation as fusion method; as in Table \ref{tab:experimental results} when dictionary explanations are used for textual entailment hypotheses and prompt initialization, prompt tuning helps increase average F1 score by 0.99\% using addition as fusion method and increase average F1 score by 1.30\% using concatenation as fusion method. 

Whether using class labels or using dictionary explanations as prompt initialization, the hypotheses obtained by prompt tuning performs better than the hypotheses converted directly from class labels or dictionary explanations by human design, which implies that prompt tuning is effective for model improvement. 
When applying hypotheses converted from class labels, the model improvement obtained by using prompt tuning is small; when applying hypotheses converted from dictionary explanations, the model improvement achieved by using prompt tuning is greater. One possible explanation is that the dictionary explanations induce longer hypotheses, allowing more parameters to be tuned during the prompt tuning phase. A good prompt template initialization can lead to better model performance.

\begin{figure*}
     \centering
     \begin{subfigure}[b]{0.45\textwidth}
         \centering
         \includegraphics[width=\textwidth]{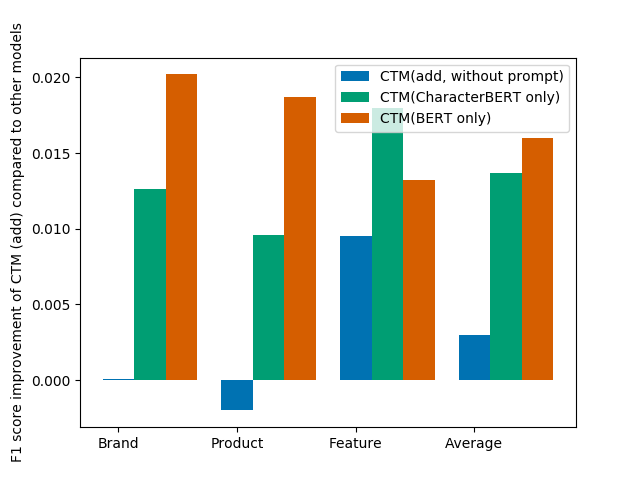}
         \caption{F1 score improvement of CTM (add) compared to other models. Class labels are used as prompt initialization and textual entailment hypotheses.}
         \label{fig:label_add}
     \end{subfigure}%
     \hfill
     \begin{subfigure}[b]{0.45\textwidth}
         \centering
         \includegraphics[width=\textwidth]{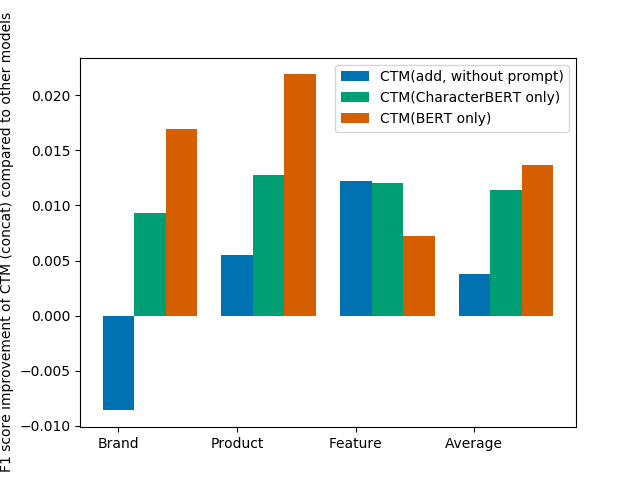}
         \caption{F1 score improvement of CTM (concat) compared to other models. Class labels are used as prompt initialization and textual entailment hypotheses.}
         \label{fig:label_concat}
     \end{subfigure}
        \caption{F1 score improvement of CTM compared to other models. Class labels are used as prompt initialization and textual entailment hypotheses.}
        \label{fig:label}
\end{figure*}

\begin{figure*}
     \centering
     \begin{subfigure}[b]{0.45\textwidth}
         \centering
         \includegraphics[width=\textwidth]{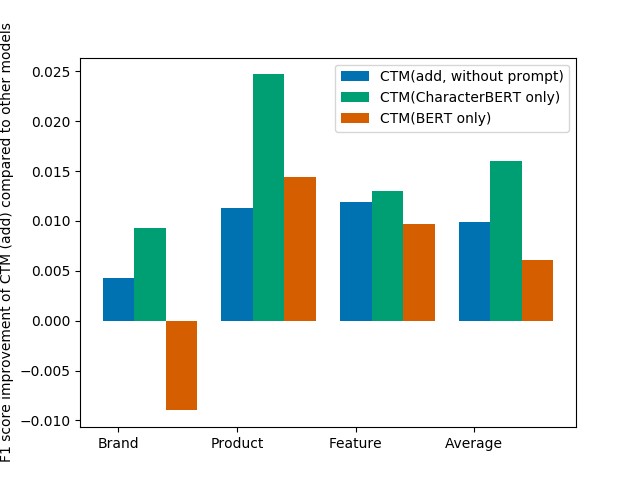}
         \caption{F1 score improvement of CTM (add) compared to other models. Dictionary explanations are used as prompt initialization and textual entailment hypotheses.}
         \label{fig:dic_add}
     \end{subfigure}%
     \hfill
     \begin{subfigure}[b]{0.45\textwidth}
         \centering
         \includegraphics[width=\textwidth]{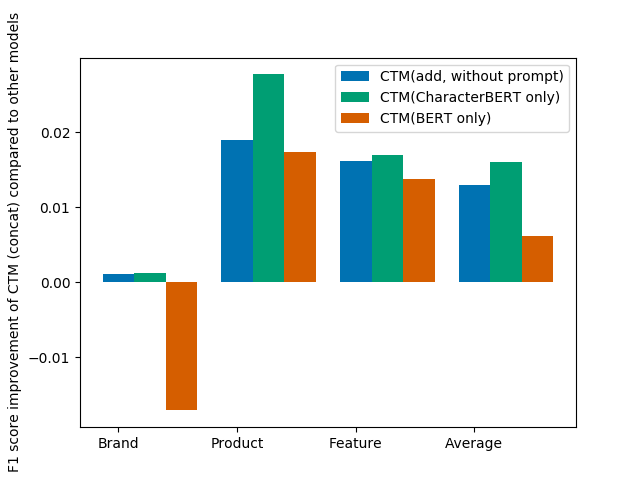}
         \caption{F1 score improvement of CTM (concat) compared to other models. Dictionary explanations are used as prompt initialization and textual entailment hypotheses.}
         \label{fig:dic_concat}
     \end{subfigure}
        \caption{F1 score improvement of CTM compared to other models. Dictionary explanations are used as prompt initialization and textual entailment hypotheses.}
        \label{fig:dict}
\end{figure*}

\noindent\textbf{Fusion Embedding (RQ4)}
In order to compare the model performance between fusion embedding and BERT / CharacterBERT embedding, we conduct experiments on our proposed CTM models and the similar models using BERT embeddings or CharacterBERT embeddings alone on hypotheses and prompt initialization converted from both labels and dictionary explanations. In addition, we build the same architecture for BERT embedding and CharacterBERT embedding, which is using a two-layer MLP classifier for embedding classification, for better comparison. 

For fusion embedding, as in Table \ref{tab:experimental results fusion embedding} when class labels are used for hypotheses and prompt initialization, fusion embedding using addition/concatenation helps improve average F1 score by 1.37\%/1.14\% compared to model without BERT and improve average F1 score by 1.60\%/1.37\% compared to model without CharacterBERT; as in Table \ref{tab:experimental results fusion embedding} when dictionary explanations are used for hypotheses and prompt initialization, fusion embedding using addition/concatenation helps improve average F1 score by 1.60\% compared to model without BERT and improve average F1 score by 0.61\% compared to model without CharacterBERT.

Fusion embedding works much better than BERT embedding or CharacterBERT embedding either using class labels or dictionary explanations as hypotheses and prompt initialization. This is expected because by using fusion embedding, we cover both word-piece-level information and character-level information, which makes the model more adaptable to product title language style and special e-commerce product title vocabulary.

\noindent\textbf{Summary}
The comparison between the entity typing and the textual entailment formulation on test set and novel entity set shows that the textual entailment model is more flexible to new entities that are not present in training set. And the different model performance achieved by applying different hypotheses to the textual entailment model confirms that the selection of hypotheses is essential to textual entailment model performance.
Hypotheses obtained using prompt tuning work better than hand-crafted hypotheses, and the proposed models with dictionary explanations as prompt initialization work better than the proposed models with class labels as prompt initialization. 
Fusion embedding has positive effects on model performance compared to BERT embedding alone and CharacterBERT embedding alone.
Fig. \ref{fig:label} and Fig. \ref{fig:dict} illustrate performance comparison of ablation studies, showing the F1 score improvement of CTM compared to other models using class labels and dictionary explanations respectively. The figures emphasize that our proposed CTM models with different initialization and fusion methods outperform almost all other models on test set.

\section{Conclusions}
\label{sec:5}
We reformulate an entity typing task in product titles into a novel textual entailment model and utilize the continuous prompt tuning method to automatically create entailment hypotheses. The textual entailment formulation provides model with the ability to better handle new entities that are not present in the training set. And the use of the continuous prompt tuning eliminates the trouble of manually designing hypotheses and creates better hypotheses to increase the model performance. We try different types of prompt templates to initialize the continuous prompt tuning method and analyze the different results obtained. Besides, we apply the fusion embedding of BERT embedding and CharacterBERT embedding to address the different language styles of product titles and special vocabularies in e-commerce. We also implement experiments to compare model performance between entity typing and textual entailment model, and conduct ablation studies on continuous prompt tuning and fusion embedding to analyze the effect of each contribution. The results show that the textual entailment formulation is more flexible to new entities than entity typing task and both continuous prompt tuning and fusion embedding have a positive effect to our model. In conclusion, our proposed CTM model has the ability to deal with new entities and the special language styles of product titles.

\section*{Acknowledgment}
This work is supported in part by NSF under grants III-1763325, III-1909323,  III-2106758, and SaTC-1930941. 

Thank Huski.ai\footnote{https://huski.ai} for sponsorship and data support.

\vspace{12pt}

\end{document}